\documentclass[twocolumn, fleqn]{svjour2}

\smartqed  
\usepackage{subcaption}
\usepackage{subfig}
\usepackage{graphicx}
\usepackage{todonotes}
\usepackage[numbers]{natbib}
\usepackage[british]{babel}
\usepackage{siunitx}
\usepackage{enumitem}
\usepackage{microtype}

\journalname{Construction Robotics}
\begin{document}

\title{Mobile Robotic Fabrication at 1:1 scale: \newline the In situ Fabricator
\thanks{
This research was supported by the Swiss National Science Foundation through the National Centre of Competence in Research (NCCR) Digital Fabrication (Agreement \#51NF40\_141853) and a Professorship Award to Jonas Buchli (Agreement \#PP00P2\_138920). \newline
The authors would like to thank Norman Hack, Nitish Kumar and Alexander N. Walzer for providing details on the Mesh Mould process. Special thanks also go to the NCCR technicians Michael Lyrenmann and Philippe Fleischmann for their technical support on IF1.}
}
\subtitle{System, Experiences and Current Developments}

\titlerunning{The In situ Fabricator} 

\author{
Markus Giftthaler \and 
Timothy Sandy  \and 
Kathrin D\"orfler \and 
Ian Brooks \and 
Mark Buckingham \and 
Gonzalo Rey \and 
Matthias Kohler \and 
Fabio Gramazio \and 
Jonas Buchli
}


\institute{   Markus Giftthaler, Timothy Sandy, Jonas Buchli \at
              ETH Z\"urich,
              Institute for Robotics and Intelligent Systems
              \\
              \email{\{mgiftthaler, tsandy, buchlij\}@ethz.ch}
              \and
              Kathrin D\"orfler, Matthias Kohler, Fabio Gramazio \at
              ETH Z\"urich, Chair of Architecture and Digital Fabrication 
              \\
              \email{\{doerfler, kohler, gramazio\}@arch.ethz.ch}
              \and 
              Ian Brooks, Gonzalo Rey \at Moog Inc.
              \email{\{grey, ibrooks\}@moog.com}
              \and 
              Mark Buckingham \at Renishaw PLC.
              \email{mark.buckingham@renishaw.com}
}


\maketitle

\begin{abstract}
This paper presents the concept of an \emph{In~situ Fabricator}, a mobile robot intended  for on-site manufacturing, assembly and digital fabrication. We present an overview of a prototype system, its capabilities, and highlight the importance of high-performance control, estimation and planning algorithms for achieving desired construction goals. Next, we detail on two architectural application scenarios: first, building a full-size undulating brick wall, which required a number of repositioning and autonomous localisation manoeuvres. Second, the Mesh Mould concrete process, which shows that an In situ Fabricator in combination with an innovative digital fabrication tool can be used to enable completely novel building technologies.
Subsequently, important limitations and disadvantages of our approach are discussed. Based on that, we identify the need for a new type of robotic actuator, which facilitates the design of novel full-scale construction robots. We provide brief insight into the development of this actuator and conclude the paper with an outlook on the next-generation In situ Fabricator, which is currently under development.
\end{abstract}


\section{Introduction}
\label{sec:introduction}
\subsection{Motivation}
\label{sec:motivation}

In the past decades, there has been significant effort to raise the degree of automation in building construction and architecture. 
\emph{Digital fabrication} promises a revolution in the construction industry and exhibits a great potential for novel architectural approaches and alternative tectonics. 

The tight integration of planning and construction allows the architect to optimise processes on multiple levels. During planning, shapes can be optimised in order to create highly differentiated forms which minimise the usage of material. For construction, novel processes are enabled which minimise material waste, increase efficiency and improve working conditions. To handle the high level of geometric- and fabrication-informed complexity in an efficient manner, using digitally controlled machinery for the construction of computer-generated forms is essential. Furthermore, by tightly integrating digital design and fabrication, the performance and aesthetics of the structures being built can be improved through continuous adaptation of the design and process in real time.

To date, digital fabrication has had the most impact in the area of off-site prefabrication in which smaller components of a building are made in a dedicated factory and then transported to the building site for final assembly (an example is the roof presented in~\cite{willmann2016robotic}). Directly on building sites, however, the level of automation is still comparably low. The final assembly of building components is heavily dominated by manual labour, which breaks the digital process chain between design and making.

Motivated by this insight, recent research goes in the direction of on-site digital fabrication, the autonomous fabrication of buildings (or building components) on the spot, generally referred to as \emph{In situ Fabrication}. Within this field, one approach addresses on-site additive manufacturing with large scale gantry systems~\cite{Khoshnevis2004,Bosscher2007}. However, the most striking disadvantage of this approach is the fact that the size of the employed machine constrains the size of the object being built. Therefore, using mobile, autonomous robots can be considered a more versatile option as it allows for the fabrication of structures significantly larger than the tool employed.

A number of attempts have already been made to develop mobile robots for on-site robotic construction. Early exploratory setups deserving explicit mentioning were the robots ``Rocco"~\cite{Andres1994rocco} and ``Bronco"~\cite{pritschow1996technological}. However, these systems were designed for relatively standardised and strictly organised production processes. ``Dimrob" was presented in~\cite{helm2012dimrob}, but several factors restricted the platform's usefulness for a wide range of building scenarios. It could only be repositioned manually and did not make use of advanced sensing and control concepts, therefore requiring the use of static support legs and effectively rendering it as a movable fixed-base robot. 
In~\cite{keating2016from}, a full-scale mobile system was shown to be capable of printing a large-scale foam structure, however also using a quasi fixed-base setup and without strict accuracy requirements. In~\cite{jokic2015minibuilders}, a self-supporting 3d printing system which can move on the printed structure was presented. For a recent, more complete survey of mobile robots in construction, we refer the reader to~\cite{ardiny2015are}.

To the best of our knowledge, to date, there is no robotic platform which fully satisfies the requirements for autonomous, mobile robotic construction at 1:1 scale. While there has been a number of research projects aiming at enabling mobile robots for on-site robotic construction, we believe that the key challenge which has prevented significant breakthroughs so far is that such machines must be able to robustly handle the unstructured nature of the building construction site. Because construction sites are constantly changing and relatively dirty and cluttered environments, it is not possibly to apply classical industrial automation approaches in controlling such systems.

This challenge poses design, engineering and research questions at many different levels. While environmental and hardware requirements (e.g. payload requirements) determine the design, shape and physical realisation of the mobile robot itself, the role of state estimation, control and planning algorithms, as well as their proper implementations, should be considered in the design of the overall system such that it can be effectively operated by a non-expert user. Finally, the system needs to be integrated into layouting systems and architectural design software in such a way that there can be a seamless interaction between design and construction.

\subsection{Contributions and Structure of this Paper}
\label{sec:contributions}
In this paper, we present the `In situ Fabricator' con\-cept: a class of mobile robot specially designed for on-site digital fabrication. First, we propose a list of basic requirements for an In situ Fabricator in Section~\ref{sec:requirements_and_definition}. Second, we present a systematic overview of the IF1, a first prototype built from off-the-shelf components, in Section~\ref{sec:in_situ_fabricator_1}. Next, we introduce the fully-integrated digital tool-chain developed for the system, spanning from digital design to the planning and control for the mobile system. In Section~\ref{sec:estimation_planning_control}, we present the planning, state estimation and control algorithms methods used for achieving the required capabilities for digital fabrication on IF1. In~Section~\ref{sec:integration_architectural}, we explain the IF1's integration into architectural design and planning Software. We highlight the capabilities of this fully-integrated system through two architectural demonstrators, in Section~\ref{sec:architectural_demonstrators}. First a dry brick wall, which demonstrates the IF1's ability to build geometrically complex shaped structures at full scale. Second, we showcase the Mesh Mould project, which shows the IF1's potential to enable completely new building processes.

Reliable, dedicated hardware plays an important role in the construction sector. The characteristics of tasks appearing in building construction differ signifi\-cant\-ly from the task spectrum that classical industrial robotics can cover today. Therefore, in Section~\ref{sec:limitations} we outline the inherent limitations of IF1 and related concepts based on commercially available industrial robotics. We list important conclusions drawn from those limitations, and highlight the development of a novel type of actuation designed specifically for the needs of full-scale robotic construction, in Section~\ref{sec:next_gen_if}. Based on that, we introduce the concept of the future In situ Fabricator (IF2), which is currently under development.

\section{Requirements and Definition of an In situ Fabricator}
\label{sec:requirements_and_definition}
Looking at a typical construction site today, one will often find a variety of machines of different sizes and with different specialised purposes. It is likely that we will see a similarly broad spectrum of different robots for specialised tasks in building construction in the future. In our research we have decided to first consider an inter\-mediately-sized class of mobile robots dedicated to a broad variety of fabrication tasks, referred to as \emph{In situ Fabricators}. We believe that such a machine could have a significant impact on building construction in the near future and would effectively demonstrate the capabilities of on-site robotic digital fabrication. In situ Fabricators are defined through the following set of requirements: \newline
\newline
\noindent \textit{Control and state estimation:}
\begin{itemize}[noitemsep, topsep=0pt]
    \item provide 1 to 5 millimetre positioning accuracy at the end effector.
    \item can operate within a local portion of the construction site. Moving obstacles, humans, and changing scenes outside of this area should not impact performance.
    \item is mobile in non-flat terrain with obstacles and challenges as found on a typical construction site.
    \item can operate with limited human intervention. The machine alone should offer the modality for achieving the overall accuracy of the building task.
\end{itemize}
\noindent \textit{Size and workspace constraints:}
\begin{itemize} [noitemsep, topsep=0pt]
    \item can reach the height of a standard wall.
    \item can fit through a standard door (in our case defined as a 80~cm wide Swiss standard door).
    \item can be loaded on a pallet/van.
\end{itemize}
\noindent \textit{Versatility and customisation:}
\begin{itemize} [noitemsep, topsep=0pt]
    \item can be equipped with different tools or end effectors to perform a wide range of building tasks.
    \item have sufficient payload to handle heavy and highly customised digital fabrication end effectors.
    \item can work in confined non-ventilated spaces.
    \item are protected against dust and water ingress.
\end{itemize}
\noindent \textit{Power supply:}
\begin{itemize} [noitemsep, topsep=0pt]
    \item can be plugged into standard mains power.
    \item has sufficient on board power for phases of construction where no external power supply is available (e.g. during transportation to and from the construction site)
\end{itemize}
\noindent \textit{Usability and integration:}
\begin{itemize}[noitemsep, topsep=0pt]
    \item can provide required information to the architectural planning and control environments, e.g. current robot location, building state, etc.
    \item provides interfaces for interaction with an operator who is not a robotics expert.
\end{itemize}
\ 

Note that we are not addressing the whole building production process chain, which would also include logistics and supply management. While this domain offers great opportunities for automation and optimisation, our work focuses on a machine intended solely or the production of the desired structure.
As such, special attention is put on creating the possibility to close the feedback loop between design and the building process through in-the loop sensing and control.

\section{In situ Fabricator 1}
\label{sec:in_situ_fabricator_1}
\begin{figure}[tb]
    \centering
    \includegraphics[width=0.99\columnwidth]{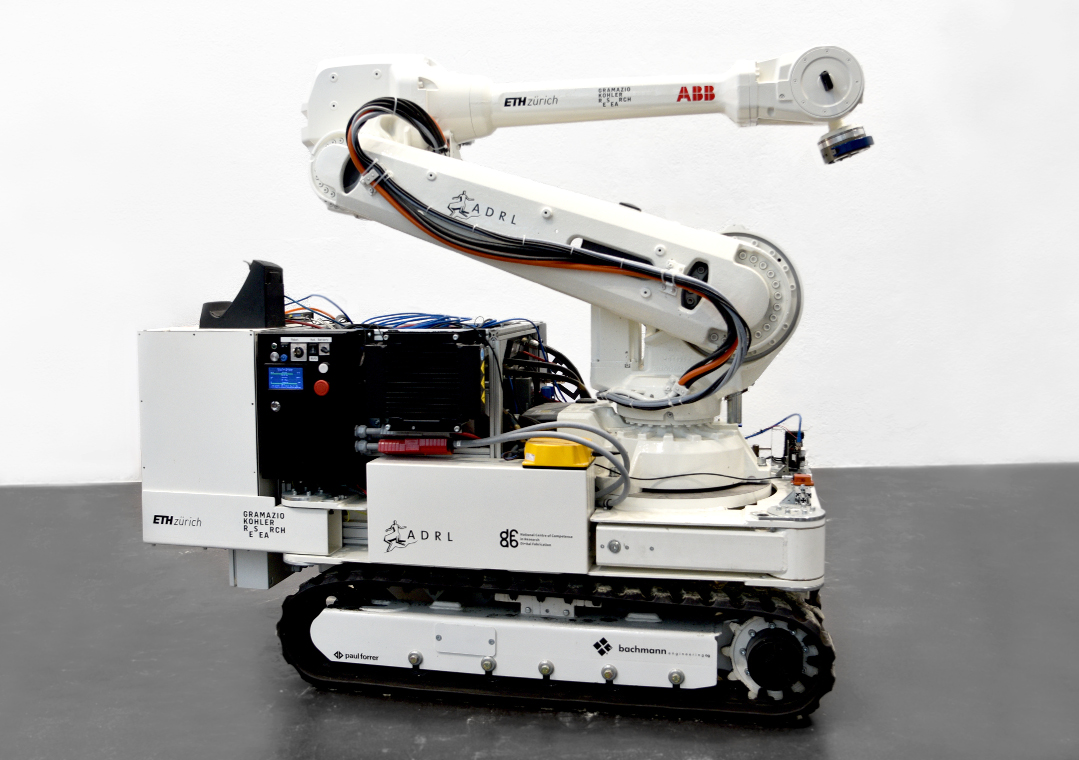}
    \caption{The In situ Fabricator 1 (without end effector).}
    \label{fig:if1}
\end{figure}
In 2014, the first prototype machine was realised, the IF1, which is shown in Figure~\ref{fig:if1}. It is partially based on existing parts from the Dimrob project~\cite{helm2012dimrob} and mostly consists of commercially available off-the-shelf components. A brief overview of the robot hardware is given in the following section.

IF1 is equipped with an ABB IRB~4600 robot arm with \SI{2.55}{\m} reach and \SI{40}{\kg} payload. The decision to use an industrial robot arm for the first prototype allowed for quick progress in providing a fully-sized mobile robot for initial research, although its limitations were already known at this point. 
All required industrial robot controller electronics from an ABB~IRC5 controller unit were fitted to the robot base in a custom, more compact, form. 
The arm is position-controlled, and a commercial control interface provided by the manufacturer allows to send reference position and velocity commands at \SI{250}{\Hz} rate. 

IF1 is electrically powered. It carries four packs of Li-Ion batteries with capacity for 3-4 hours of autonomous operation at average machine load without being plugged into mains power. The robot features an on-board charging system and a power conversion system offering currents between 5-\SI{48}{\V} DC and 230-\SI{400}{\V} AC at \SI{50}{\Hz}.

The robot also carries a custom on-board hydraulic system, which is used to power its tracks through hydraulic motors, but can also provide hydraulic power to the tool mounted at the end effector. Its core components are a compact AGNI DC electric motor attached to a pump delivering hydraulic pressure at \SI{150}{\bar}. The hydraulic system is designed such that the tracks can be driven both with manual levers or through automatic operation, in which case the flow to the tracks' hydraulic motors is controlled by proportional valves. The IF1 can achieve a maximum driving speed of 5~km/h at a total weight of 1.4~tons.

Depending on the desired task, IF1 can be equipped with additional exteroceptive sensors. All sensors and actuators are driven by an on-board computer system which runs a hard-real-time enabled version of Linux with the Xenomai kernel-patch~\cite{xenomai}. The main on-board computer unit features an Intel~i3-3220T processor with \SI{2.8}{\GHz} and 4~GB RAM, which is sufficient to run basic state estimation, planning and control algorithms. Computationally more intensive tasks are run externally, with wireless communication provided through ROS~\cite{quigley2009ros} or a custom real-time enabled TCP/IP implementation. 

The standard mounting flange of the industrial arm and a general set of power and data connections are provided at the end-effector to allow for the attachment of a wide range of tools. IF1 also provides various mounting points for temporary (complementary) equipment such as vacuum pumps or welding equipment.
%

\section{State Estimation, Planning and Control}
\label{sec:estimation_planning_control}
For enabling autonomous localisation, driving and building, we have implemented a mix of well-established as well as novel algorithms for state-estimation, planning and control. The methods described in the following sections have proven to reach high positioning accuracies over the course of long building processes and many robot repositioning manoeuvres without reliance on external reference systems. 

\subsection{Sensing and State Estimation}
In order for a robot to build structures on the construction site with high accuracy, it needs to be able to track the position of the tool it is using with respect to some fixed reference frame. This section describes the sensing system developed for IF1. These developments are broken up into three main functional parts: robot localisation within the construction site, alignment between the sensing reference frame and the CAD model, and feedback of the building accuracy during construction.

While there is an extensive body of research from the robotics community in localisation~\cite{bonin2008visual}, motion tracking~\cite{scaramuzza2011visual}, and mapping~\cite{thrun2002robotic} for mobile robots, these systems do not directly translate to the construction site. One main reason for this is that most robotics applications do not consider the millimetre-scale relative positioning accuracy required for building tasks. Another reason is that most applications do not consider any prior information about the environment in the sensing system, while there is an abundance of prior information for building construction, in the form of the CAD model or other plan data. Our work in sensing for IF has therefore focused on tailoring existing sensing solutions from the robotics community to the application of on-site building construction.

\subsubsection{Localisation}
The most basic function of an In situ Fabricator's sensing system is to localise the end-effector of the robot with respect to a fixed reference frame. In conventional industrial robots, this is easily achieved using the rotary encoders in the robot's joints since the robot is sufficiently stiff and rigidly attached to the ground. For a mobile robot, however, exteroceptive sensing is required to ensure zero-drift pose estimates. A strategy typically used for such a system is to track a known point on the robot with respect to a static sensor system (e.g. a Vicon motion tracking system or a Hilti Total Station). While these systems can provide high-accuracy positioning data with minimal integration effort, they can be prohibitively expensive and take considerable initial setup time and effort. Furthermore, the measurement frequency and delay is often not optimized for mobile applications. Alternatively, sensors can be mounted directly on the robot and used to locate visual references in the robot's workspace. While these solutions are much lower cost, they typically require significantly more integration effort, as the sensor information must be heavily processed to extract the information required for localisation (e.g. image processing to extract the local visual features). This strategy is pursued for IF as to avoid the presence of visibility constraints from an external measurement system and because we believe this sensing modality can be more easily expanded to feed back additional pieces of information to inform the remainder of the building process (as in  Section~\ref{sec:feedbackaccuracy}).

\begin{figure}
    \centering
    \includegraphics[width=0.98\columnwidth]{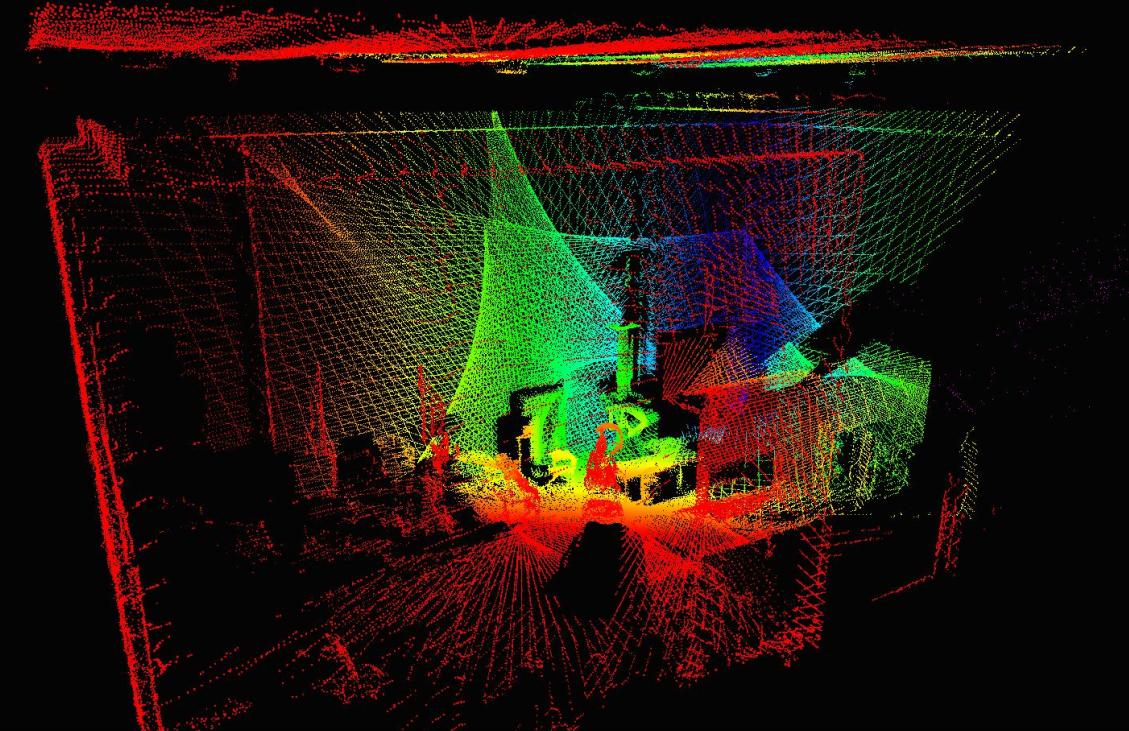}
    \caption{Point cloud of our lab captured by IF1.}
    \label{fig:cube_point_cloud}
\end{figure}

We have developed two separate sensing systems for use on IF1, each supporting one of the application examples presented in Section~\ref{sec:architectural_demonstrators}. For the brick-laying experiments, we mounted an off-the-shelf laser-range-finder (LRF) on the end-effector of IF1. By executing sweeping motions with the wrist of the arm, we could build 3D point clouds of the robot's surroundings (Figure~\ref{fig:cube_point_cloud}). These point clouds were then registered versus an initial point cloud to infer the robot's relative motion (see \cite{doerfler2016automation} for implementation details). The main shortcoming of this method is that it assumes that the majority of the robot's environment remains unchanged during construction, which is a bad assumption since construction sites are constantly changing environments. Subsequent efforts therefore focused on sensing modalities which allow the robot to localise using only features located directly in the vicinity of its workpiece, see~\cite{sandy16autonomous}.

\begin{figure}
    \centering
    \includegraphics[width=0.98\columnwidth]{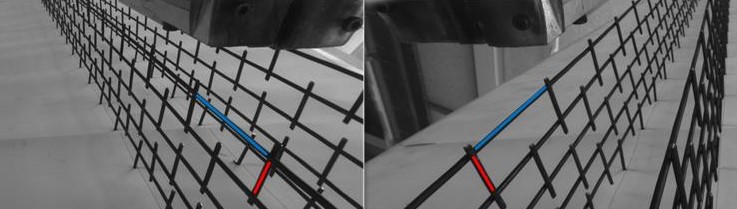}
    \caption{Stereo image pair taken with the cameras on the Mesh Mould toolhead during fabrication. The two detected wire segments are highlighted in blue and red in each image. The next segment of the mesh was welded to these segments.}
    \label{fig:line_detection}
\end{figure}

For later experiments, we switched to camera-based sensing solutions. We find that the main advantage of using cameras over LRFs is their adaptability to the application. While the absolute accuracy of LRF measurements is determined by the sensor's internal hardware, camera setups give the user the flexibility to `tune' the system's accuracy to the application by changing the resolution, lens, or position of the cameras. For the Mesh Mould project presented in Section~\ref{sec:mesh_mould}, we use the same model of camera to perform two very different sensing tasks by simply configuring and positioning the cameras differently. For localisation, any number of cameras fixed to the base of the robot observe AprilTag~\cite{olson2011tags} fiducial markers to localise relative to a calibrated map of the tag positions in the workspace. For localisation of the wire mesh and collision avoidance of the tool with the mesh, a stereo pair of cameras use line detection and matching to locate the next wire to be processed during mesh buildup. Figure~\ref{fig:line_detection} shows a stereo image pair from this system with the detected wires highlighted. With this sensing system, it is easy to see how the same cameras can be tailored to very different sensing tasks simply by modifying their configuration and the subsequent image processing. For this reason, we feel that cameras are the sensing modality with the highest potential for on-site robotic building construction.
\begin{figure}
    \centering
    \includegraphics[width=0.98\columnwidth,trim={7cm 5cm 1cm 3cm},clip]{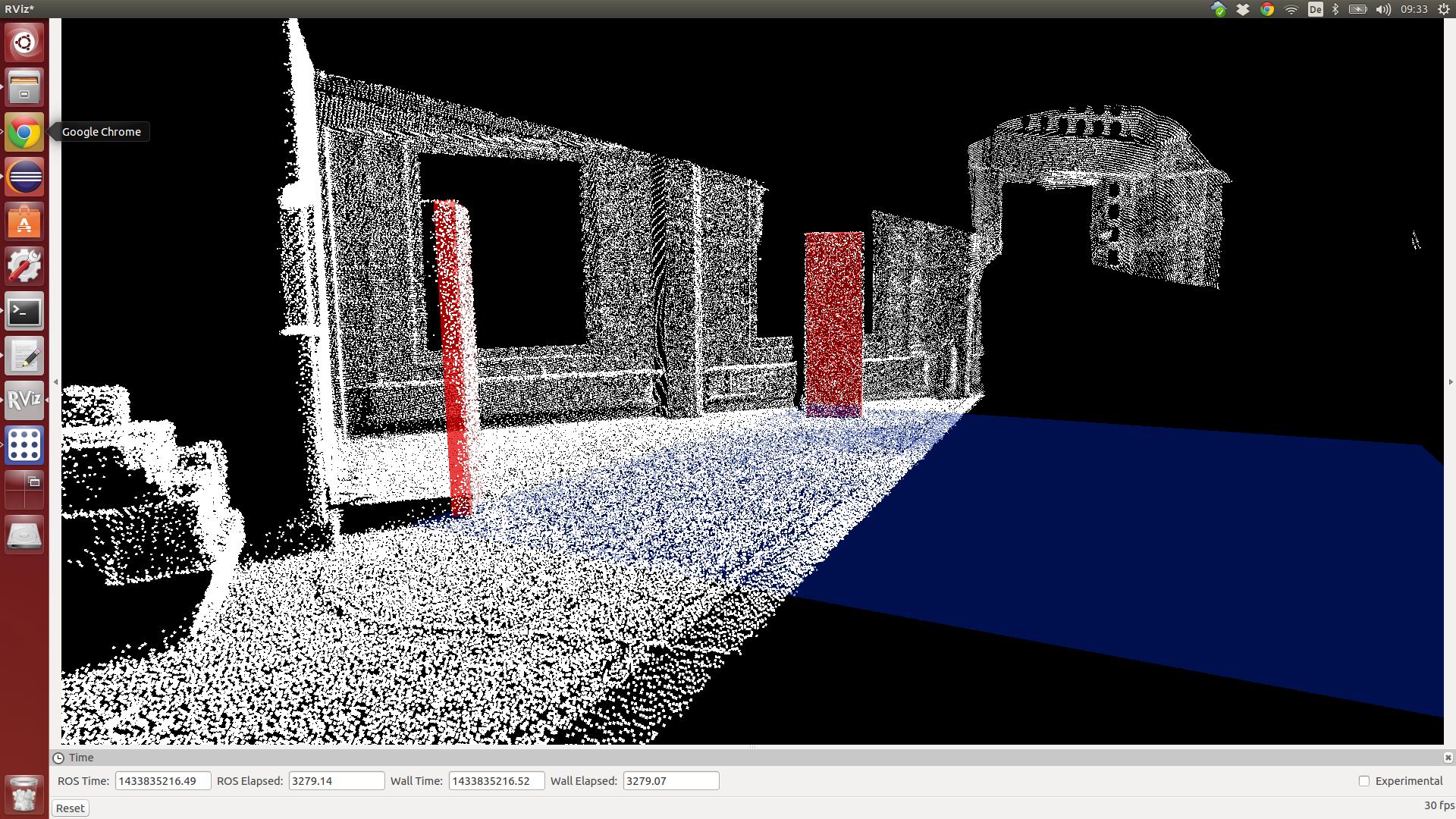}
    \caption{Point cloud of the building site, showing the registered positions of geometric models of the attachments pillars (red) and floor (blue) of the CAD model.}
    \label{fig:pillar_matching_after}
\end{figure}

\subsubsection{Alignment with the CAD Model}
In most architectural applications, localising the tool with respect to some general fixed global reference frame is not enough. Any structure built needs to be attached to an existing element of the construction site and must be be located with sufficient precision that the overall accuracy of the construction is ensured. For in situ fabrication, this can be achieved by aligning the sensing reference frame with the reference frame of the CAD model of the structure being built. This requires determining the position of either the external sensing system or the visual features used for localisation in the CAD model. This alignment step can typically be done just once before building. During this alignment step, key interfaces on the construction site, to which the structure being build must attach, can be identified and their positions fed back to update the design of the structure to accommodate any inaccuracies which may be present.

In the IF1 brick laying work (Section~\ref{sec:brick_wall}), localisation is performed relative to a reference scan of the construction site taken before starting to build. The reference scan is aligned with the CAD model of the structure by registering features of the constructions site to which the wall was anchored (Figure~\ref{fig:pillar_matching_after}). In this way, the robot is not only able to localise relative to the CAD model, but we are able to adjust the parametric design of the wall to match the true positions of the pillars within the construction site. For the Mesh Mould project (Section~\ref{sec:mesh_mould}) we need to align our map of the visual fiducial poses to the CAD model of the wall. This is done simply by using the mesh detection cameras to locate the first layer of the mesh, which is built by hand and attached to the supporting wall before building. Once this layer is located, the tags seen from that robot position can be used to align the tag map with the first layer of the mesh in the CAD model.

\subsubsection{Feedback of Building Accuracy}
\label{sec:feedbackaccuracy}
We believe that, in order to realise the full potential of robotic on-site construction technology, the robot used should be able to feed back data about the progress of the building project during construction. In this way, building inaccuracies and changing conditions in the construction site can be compensated for during construction. This is especially crucial in building processes where material does not behave predictably after processing. This is the case in the Mesh Mould project. As the wire mesh is constructed, internal tension between the wires tends to pull the mesh away from the position in which it was welded. The direction and amount of deflection is difficult to model, therefore real-time feedback of the material behaviour is required to ensure accurate construction. By using the two coupled measurement systems described, however, IF can determine the shape of the last wire welded to the mesh in the global reference frame, therefore observing how well it matches the initial design. In the presence of significant errors in the wire contour, the building plan for the next layers of the mesh can be adjusted to compensate for the error and effectively pull the mesh back into alignment with the CAD model. It should be noted that this functionality is a natural extension of the camera sensing system since various custom features can be extracted from images relatively easily. It would be very difficult, however, to do this compensation if a commercial off-board sensing systems was used, since it would not be capable of detecting the contour of the wire mesh.
\begin{figure*}[!t]
\centering
\setlength{\tabcolsep}{-1pt} 
\begin{tabular}{ccccc} 
    \includegraphics[width = 0.196\textwidth]{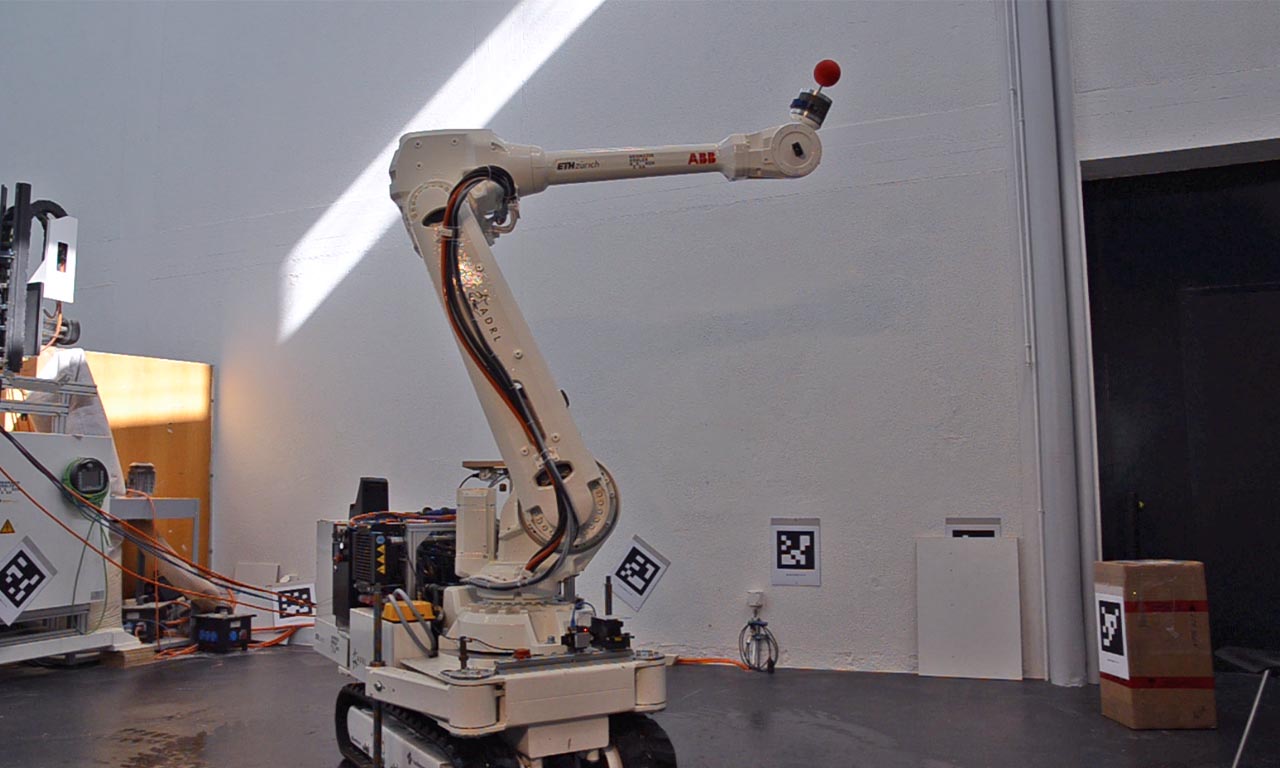} \hfill
&   \includegraphics[width = 0.196\textwidth]{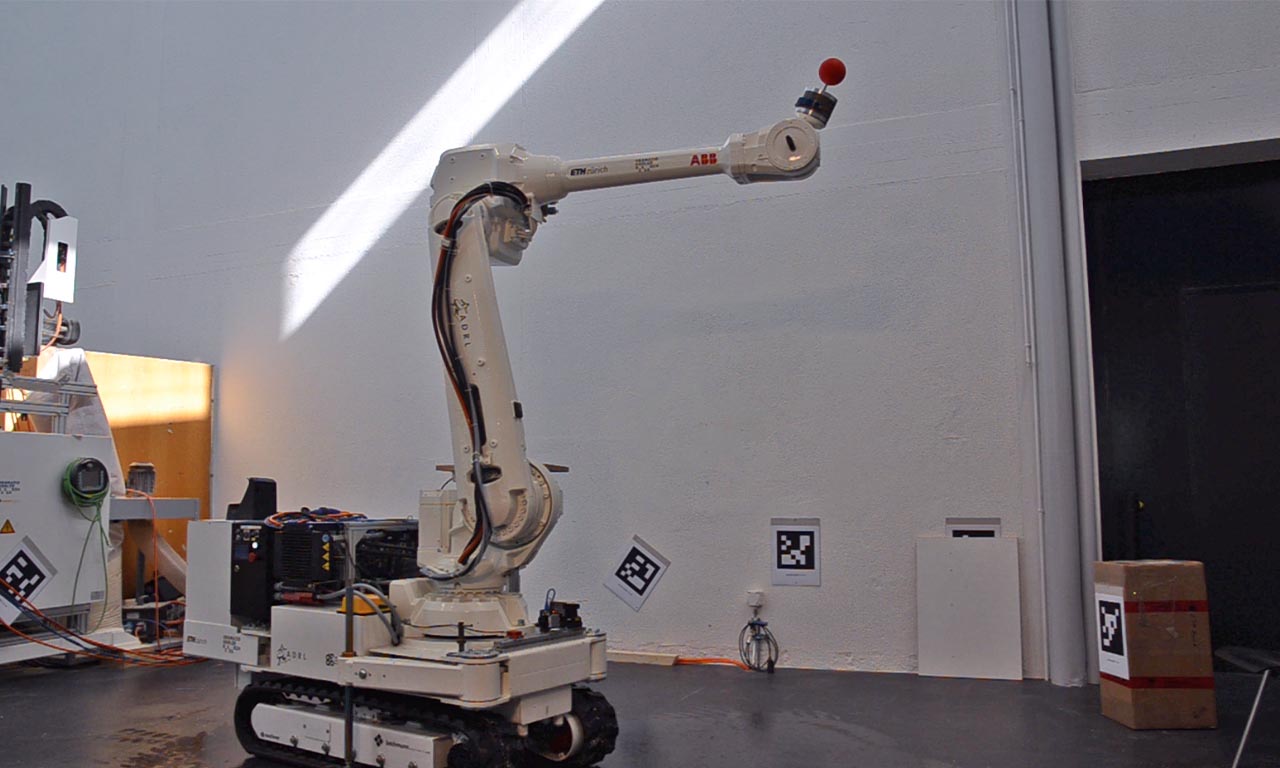} \hfill
&   \includegraphics[width = 0.196\textwidth]{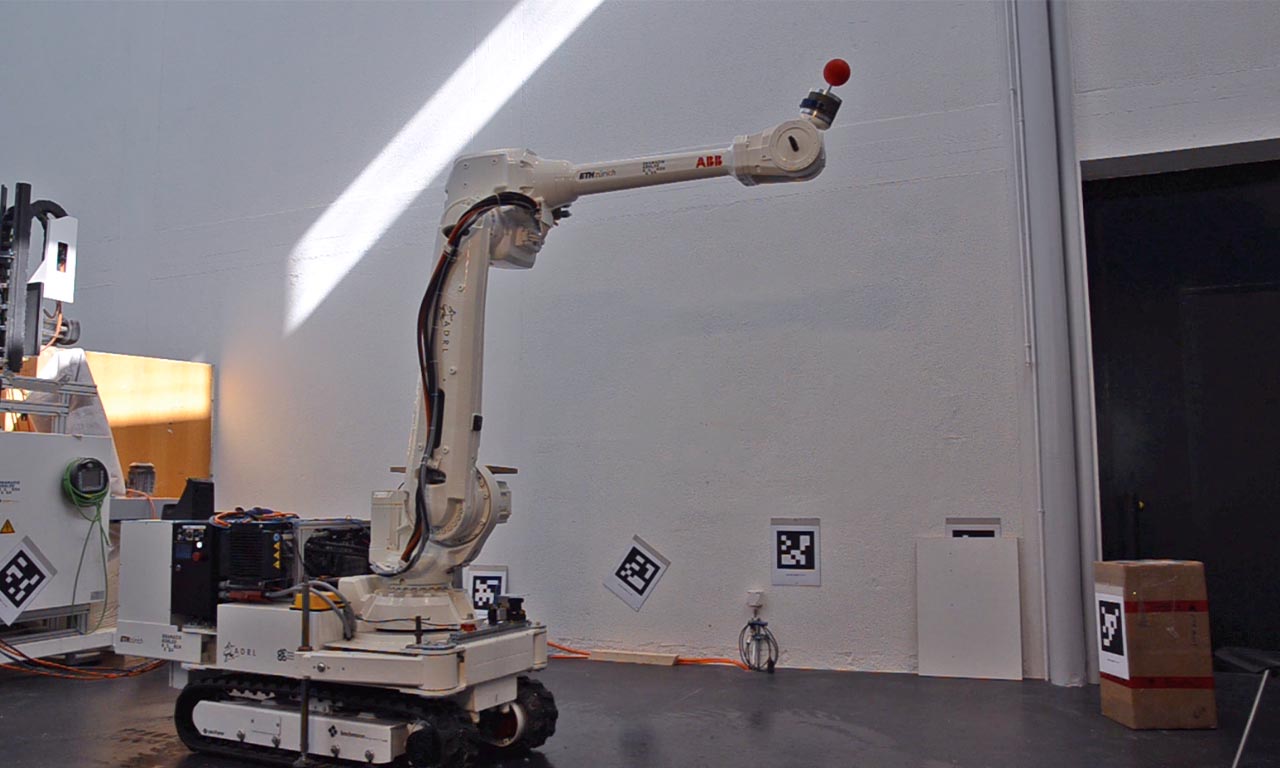} \hfill
&   \includegraphics[width = 0.196\textwidth]{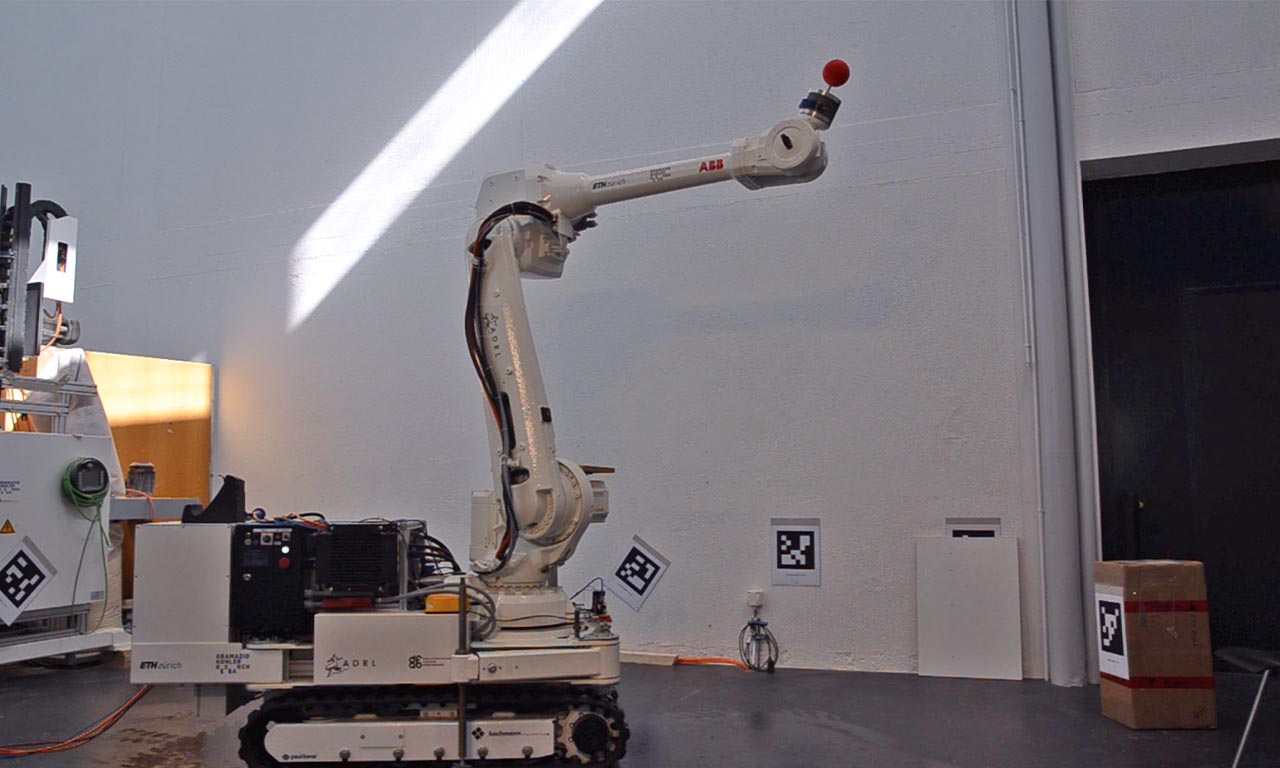} \hfill
&   \includegraphics[width = 0.196\textwidth]{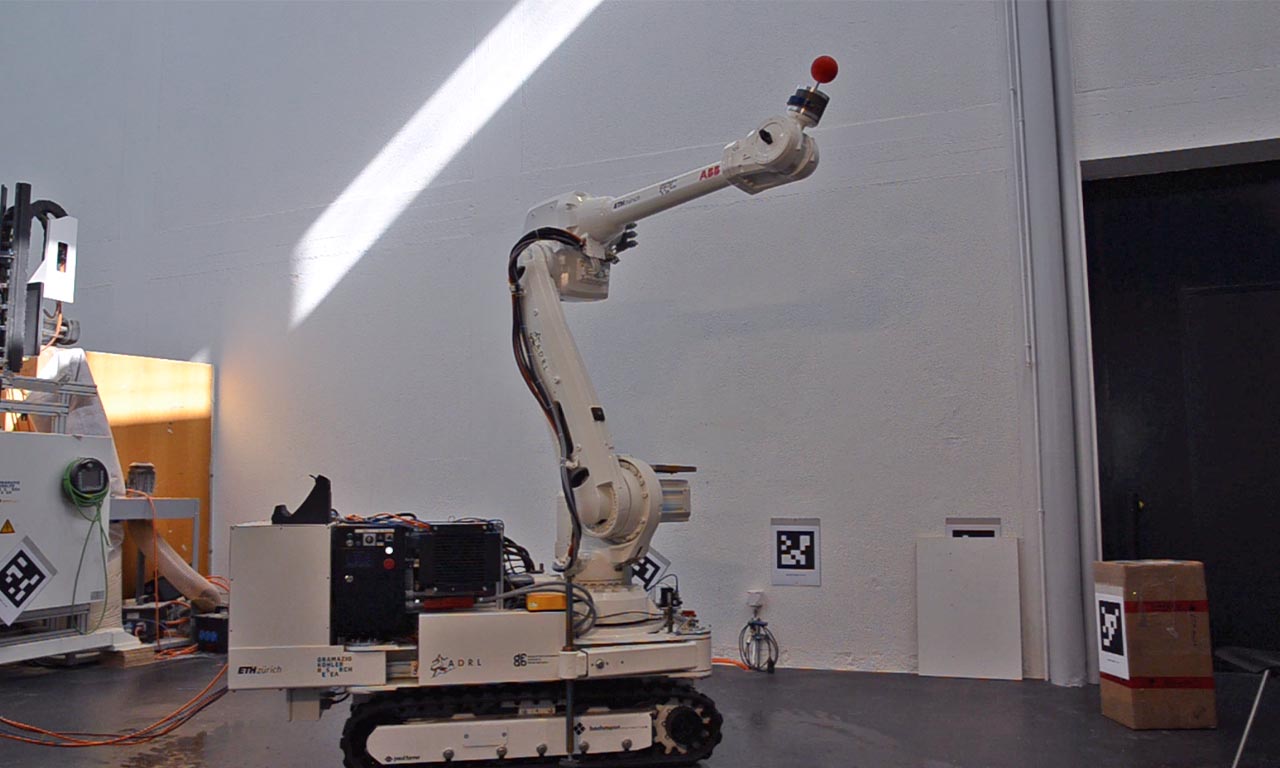} \hfill
\end{tabular}
\caption{Snapshots from a motion sequence with an end effector position constraint executed on IF1 using Constrained Sequential Linear Quadratic Optimal Control in a Model Predictive Control fashion. The task is to reposition and reorient the base while keeping the end effector at a constant position.}
\label{fig:constrained_motion}
\end{figure*}

\subsection{Planning and Control}

Generally speaking, we approach the planning and control problem for In situ Fabricators through \emph{Optimal Control}. Optimal Control solves the problem of finding a control policy for a dynamic system such that a predefined criterion of optimality is achieved. It can be used to compute either open-loop trajectories, which is the domain of trajectory optimisation, to compute feedback laws which stabilise given trajectories, which is pure feedback control, or both at the same time. 
By solving a corresponding mathematical optimisation problem, we find optimal trajectories and/or control laws that steer the system to a desired pose while minimising some cost function and respecting constraints at the same time. 
Difficulties arising in planning and control for mobile construction robots are obstacles, inherent motion constraints (for example non-holonomic constraints due to wheels or tracks), motion with contact and interaction forces and accumulated model uncertainties. The latter is an important issue for IF1, as tracked locomotion on imperfect ground can not be modelled with high accuracy. 
In contrast to classical, sam\-pling based planning algorithms like Rapidly Exploring Random Trees and Probabilistic Roadmap methods (see~\cite{lavalle2006planning} for an overview), or the integrated kinematic planners typically supplied with industrial robots, many Optimal Control algorithms can handle some of these difficulties with reasonable complexity.

For IF1, we have integrated different control and planning approaches, which allow us to consider base- and arm motion either separately or jointly as a whole-body problem.
Coordinated whole-body motions with non-holo\-nomic base constraints and holonomic operational-space constraints (e.g. tool position constraints) are generated using Constrained Sequential Linear Quadratic (SLQ) Optimal Control. Although being an iterative Optimal Control algorithm, it is computationally highly efficient, as it features linear time complexity $O(n)$. Feedforward trajectories and feedback are optimised simultaneously, which generalises the control policy in the vicinity of the nominal, optimal trajectory. The resulting feedback gains are compliant with non-holonomic constraints.
On IF1, we run this algorithm in a Model Predictive Control (MPC) fashion at up to 100~Hz update rate, where the feedback loop is closed through an on-board visual-inertial state estimator. This allows us to achieve robust positioning despite the presence of model uncertainties and external perturbations.
Figure~\ref{fig:constrained_motion} shows snapshots from a motion sequence with an end effector position constraint executed on IF1 using Constrained SLQ in an MPC fashion. More details about SLQ MPC on IF1 are provided in~\cite{giftthaler2017efficient}. 

When performing sequential building tasks, as for example demonstrated in~\cite{sandy16autonomous}, or when more obstacles are present, we separate the base- and arm control problem and move base and arm sequentially. In this case, we use a constrained version of the stochastic planner STOMP~\cite{kalakrishnan2011stomp} for trajectory optimisation. For trajectory following, one can still apply the constrained feedback gains obtained from SLQ. 

For manipulation, we combine individually planned sequences of arm/base motion with a library of task-specific, pre-programmed manipulation-primitives (e.g. picking up a brick from the brick feeder or moving a joint at constant velocity). This combination has has proven sufficient for a number of building tasks.

Note that our approach can typically handle a moderate number of obstacles easily and reliably. However, at the current stage, we cannot handle heavily cluttered environments with a large number of possibly intersecting or dynamically changing obstacles. This would lead to strongly non-convex and ill-posed optimisation problems which cannot be treated in a classical Optimal Control setting. The combination of our Optimal Control framework with higher-level planners which are able to negotiate heavily cluttered environments is part of our future research.

\section{Integration into Architectural Design and Planning Software}
\label{sec:integration_architectural}
A major interest in the development of the In situ Fabricator is to tightly integrate its functions and capabilities into an architectural planning framework, in order to make its features directly available for architects and designers. Eventually we are aiming to see the generation and rationalisation of shapes to be directly influenced by the specific logic of making -- in this case, next to the choice of a material and assembly system, this is the feature of mobility and the extended workspace of the mobile robot.

To fully exploit the design-related potentials of using such a robot for fabrication, it is essential to make use not only of the manipulation skills of this robot, but to also use the possibility to feed back its sensing data into the design environment. This allows the system to guide and inform a running fabrication process such as to be able to detect and react upon unforeseen assembly tolerances and process-related uncertainties. 
Furthermore, the system can base immediate design decisions on the information extracted from sensors, allowing a high level of flexibility, autonomy and control in fabricating an architectural artifact.

Motivated by this, the high level planning of fabrication tasks, such as the sequencing of the mobile robot's positions and fabrication procedures and computing the arm, base and end effector positioning commands, is implemented within an architectural planning tool, in our case Grasshopper Rhinoceros~\cite{rhinogh}. A TCP/IP plugin allows for the online control the robot's arm and base, and gives access to the robot's state estimator, planning routines and movement primitives. This approach is also detailed in~\cite{doerfler2016automation, kumar2017design}. Generally speaking, the robot's setup is designed to allow for feedback loops at multiple levels of the system: All time-sensitive tasks are executed by control loops running on the robot's low-level computer while the control loop over the overall building process is closed via the architectural planning tool.

\section{Architectural Demonstrators and Examples}
\label{sec:architectural_demonstrators}
The main drivers for the development of the IF1's functionality and software framework were the architectural demonstrators shown in this section.
The challenge of fabricating multiple architectural prototypes with an increasing level of complexity was specifically chosen to gradually advance the generic features of the robot.
The realisation of these demonstrators was significant for evaluating chosen methods and to learn what is necessary -- from both the robotics and the architecture point of view -- to enable automated material deposition and assembly processes in an unstructured, cluttered, and ever changing environment such as a construction site. 
To date, this enables us to build customised, geometrically complex structures accurately over the course of the entire building space.

\subsection{Undulating Brick Wall}
\label{sec:brick_wall}
\begin{figure}[tb]
    \centering
    \includegraphics[width=0.98\columnwidth]{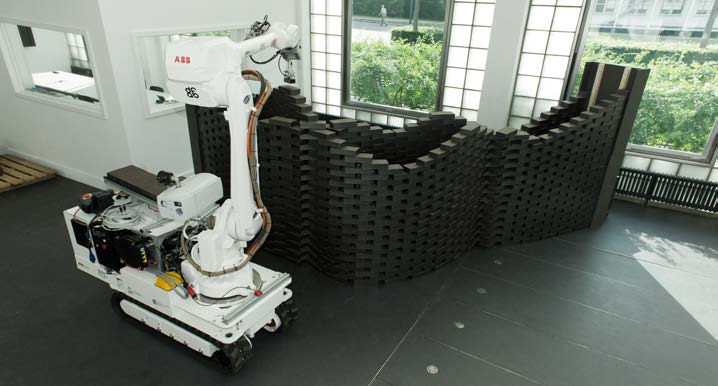}
    \caption{IF1 building the undulating brick wall.}
    \label{fig:brick_wall}
\end{figure}
\begin{figure}[tb]
    \centering
    \includegraphics[width=0.99\columnwidth]{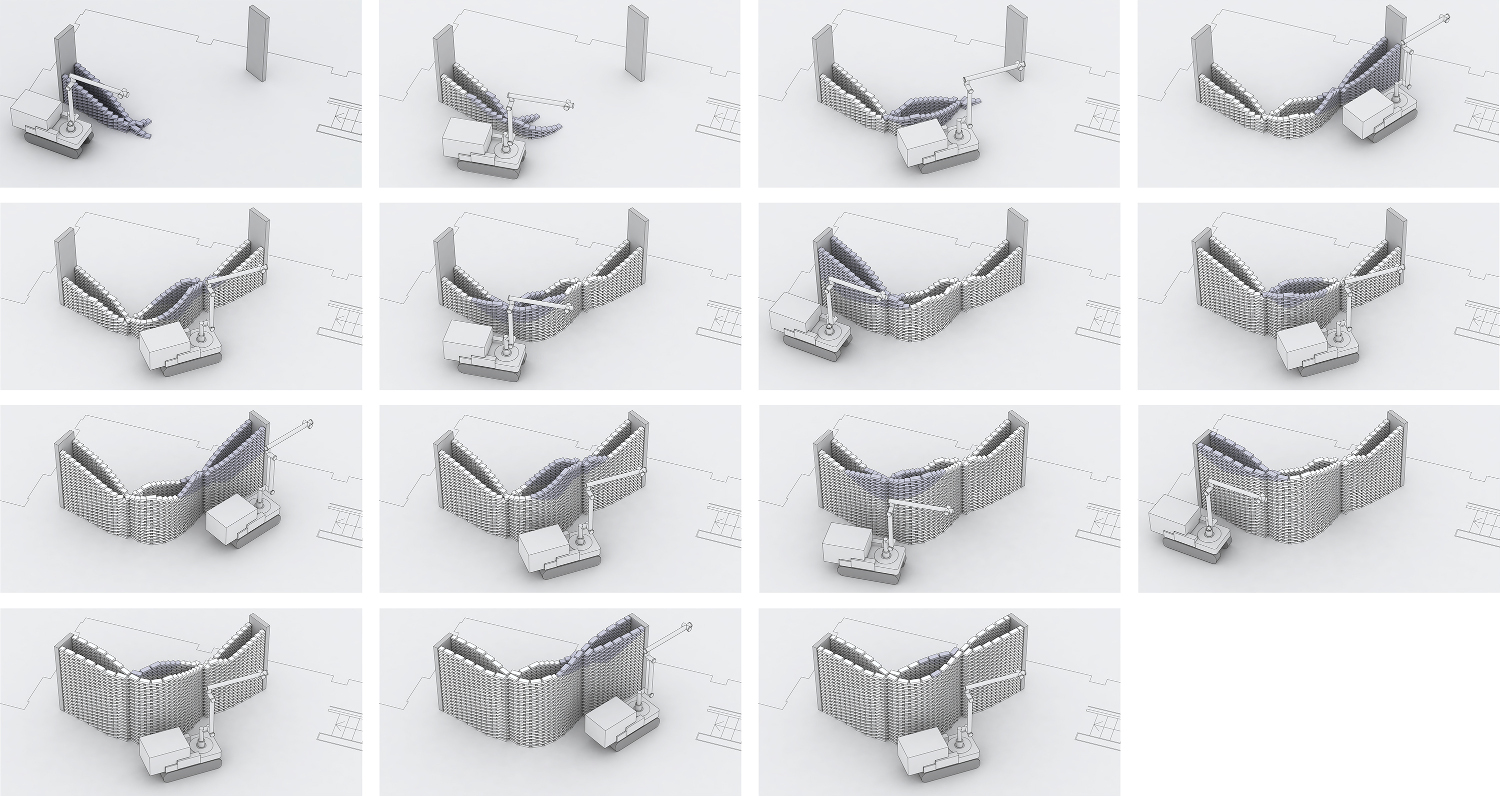}
    \caption{Visualisation of a possible building sequence for the double-leaf brick wall shown in Figure~\ref{fig:brick_wall}. The wall is \SI{6.5}{\metre} long and \SI{2}{\metre} high, consists of 1600 bricks and is fabricated from 15 different base positions.}
    \label{fig:brickwall_building_sequence}
\end{figure}
The first architectural scale demonstration with IF1 was the semi-autonomous fabrication of a continuous dry-stacked, undulating brick wall (Figure~\ref{fig:brick_wall}) in a laboratory environment which was set-up to mimic a construction site. The material system -- consisting of discrete building elements and a simple assembly logic -- allowed us to subdivide the sequential building process of the entire wall into discrete production steps from subsequent robot locations (Figure~\ref{fig:brickwall_building_sequence}).

In this experiment, it played a key role to align the CAD model of the building site with the true positions of key features of the working environment, extrapolated from the initial 3D scan of the surrounding captured by the robot's sensor system. This allowed us to adapt the ideal dimensions of the wall's parametric geometry model to the true dimensions of the construction site before actually starting the fabrication.

The building process itself consisted of iterative steps of moving the robot to positions along the wall, localising the robot's base pose and building a patch of bricks reachable within the workspace of the robot. The global localisation and brick placement errors did not accumulate over the course of the building process, first because of the initial alignment of the true positions of the attachment points to the building plan, and second, every point cloud for localisation captured from a new location was always registered against the same initial reference scan. Therefore, the designed double-leaf brick wall -- requiring the robot to be repositioned 14 times -- was successfully constructed semi-autonomously with a maximal assembly error of \SI{7}{\mm} over the entire workpiece. While the bricklaying was done autonomously, feeding the robot with bricks was accomplished manually. 

\subsection{Mesh Mould}
\label{sec:mesh_mould}
\begin{figure}[tb]
\centering
   \begin{subfigure}[b]{0.49\columnwidth}
   \includegraphics[width=0.9\columnwidth]{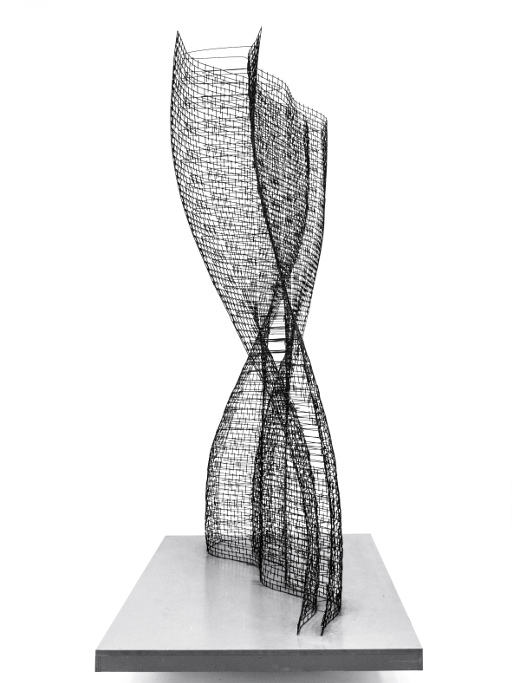}
   \caption{Empty mesh.}
   \label{fig:empty_mesh} 
\end{subfigure}
\begin{subfigure}[b]{0.49\columnwidth}
   \includegraphics[width=0.9\columnwidth]{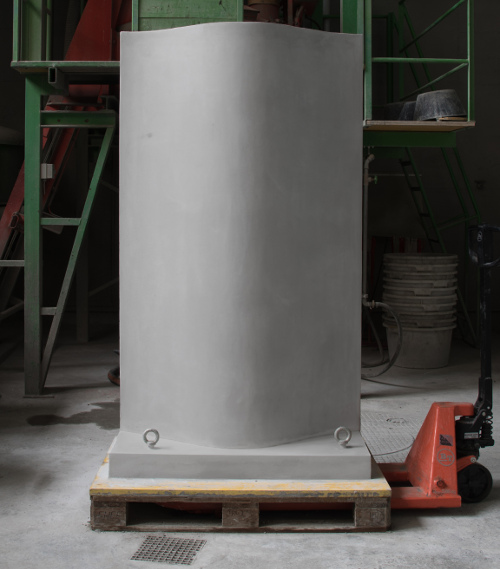}
   \caption{Filled mesh.}
   \label{fig:filled_mesh}
\end{subfigure}
\caption{Mesh Mould demonstrators built by IF1.}
\label{fig:mesh_examples}
\end{figure}
The second demonstrator combines the novel construction technique Mesh Mould~\cite{hack2015meshmould, kumar2017design} with the use of IF1. The main objective of Mesh Mould is the bespoke fabrication of free-form steel meshes which form both mould and reinforcement to enable a waste-free production of customised reinforced concrete wall structures (see Figure~\ref{fig:mesh_examples}). This fabrication method is an ideal test-bed for showing the possibility of a continuous construction process fabricated by a mobile robot. The possibility to bend and weld these meshes directly on site offers a multitude of advantages: the integrated vision feedback system allows to react to material tolerances during fabrication on the spot: It allows to negotiate between true measurements of the structure during build-up and a required target shape based on the planning data right when it is needed. Production sequences can radically be redefined: Structures to be built do  not have to be discretised into separate building components due to size limits for transportation, but can rather be redefined in accordance with the fabrication logics of the chosen material system and the mobile machinery.

\begin{figure}[tb]
    \centering
    \includegraphics[width=0.99\columnwidth]{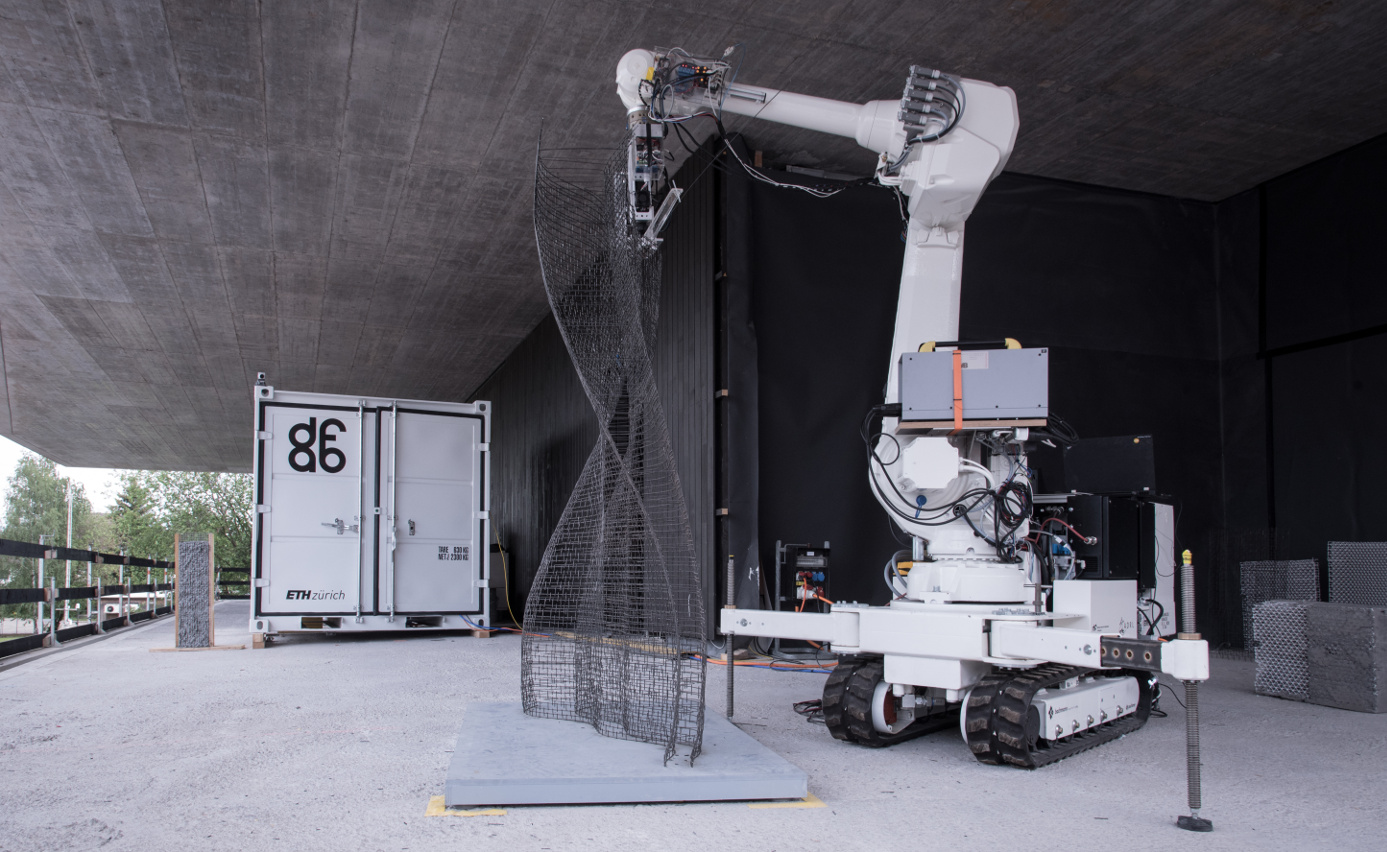}
    \caption{IF1 building a small doubly-curved metal mesh at the exploratory construction site NEST.}
    \label{fig:if_on_nest}
\end{figure}

The integration of the robotised Mesh Mould end effector for bending and welding steel wires into the architectural control and simulation framework of IF1 constituted a major part of the efforts in implementation. Once completed, the whole system was tested for the first time on a floor slab of NEST\footnote{\label{footnote_nest}https://www.empa.ch/web/nest} -- an exploratory construction site at EMPA in Z\"urich. The fabrication of undulated doubly curved Mesh Mould elements directly on the construction site served to assess the robustness of the system and requirements in the logistics of performing the fabrication process in situ (see Figure~\ref{fig:if_on_nest}). 

\begin{figure}[tb]
    \centering
    \includegraphics[width=0.99\columnwidth]{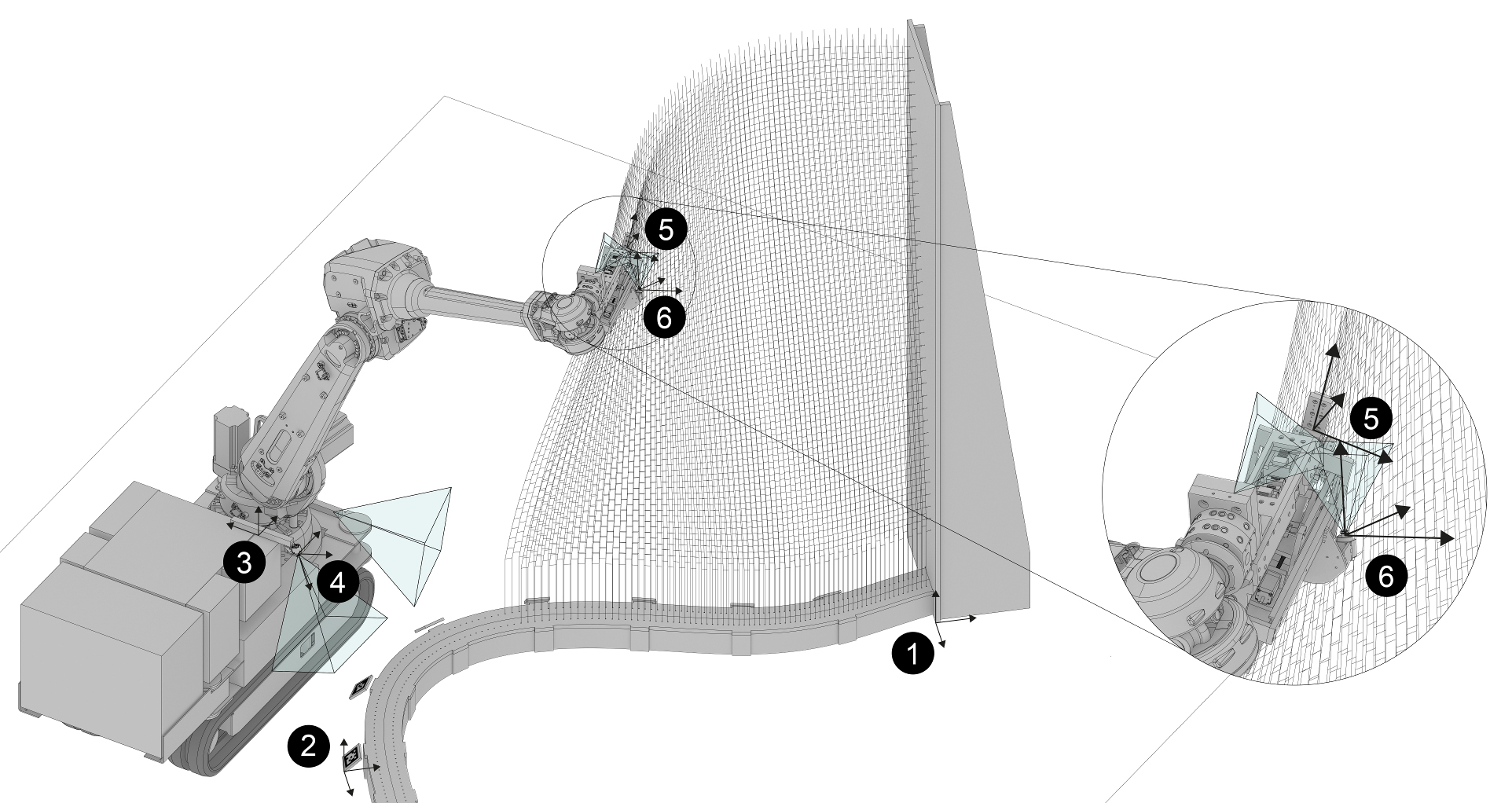}
    \caption{Vision system and frame definition for the Mesh Mould wall fabrication process: \textbf{1} world frame, \textbf{2} tag frame, \textbf{3} robot frame, \textbf{4} base camera frame, \textbf{5} end effector frame, \textbf{6} end effector camera frame.}
    \label{fig:mesh_mould_process}
\end{figure}

For the Mesh Mould project, IF1 was equipped with both a global and local state estimator (see Figure~\ref{fig:mesh_mould_process}). The global pose estimation of the robot's origin using artificial landmarks enables automated localisation during repositioning procedures. Further, the integration of perception at the end effector is required for correcting the end effector pose in case of detecting accumulative fabrication errors and mesh deformations.

In 2017, IF1 will be integrated into a larger building project. It is planned to fabricate a fully load bearing \SI{14}{\metre} long steel reinforced concrete wall at the ground level of the NEST unit realized by the Swiss National Competence Center of Research in Digital Fabrication\footnote{http://www.dfab.ch}.
%

\section{Limitations and lessons learned from IF1 - why classical industrial arms are a poor choice for mobile building construction robots}
\label{sec:limitations}
IF1 by design exhibits a number of drawbacks. Importantly, these shortcomings are not specific to this particular robot, but are rather inherent to the relevant off-the shelf technology existing to date and being available to research in our field. Some of the predominant issues are summarised in the following section.

Standard serial-chain industrial manipulators often make use of heavy-duty electric motors and gearboxes. The joints and links are designed for maximising stiffness, which is essential for reaching high positioning accuracy using traditional robot control approaches. A consequence resulting from that design strategy is a relatively low payload to weight ratio (PWR). For example, our ABB IRB 4600 arm offers a PWR of \SI{40}{\kg}:\SI{440}{\kg}. Additional downsides to the weight of industrial robots are the need for a heavy base to ensure that the robot cannot tip over, the difficulty transporting the robot, and the added safety risks of operating such a large system. At 1.4~tons, IF1 is already too heavy to access some standard building environments.

Purely position-controlled robotic arms are by design ill-suited for many construction tasks. For advanced manipulation tasks taking place beyond perfect conditions, such as on-site assembly of structures, drilling, coring or chiseling, being able to control the interaction forces between tool and workpiece is essential. While adding a multi-DoF force-torque sensor at the end effector appears to be workaround and certainly gives more flexibility to the setup, it remains a sub-optimal design choice. While a detailed discussion of this issue is beyond the scope of this paper, it is a well established result that such an arrangement (non-collocated sensing and control) has non-ideal control theoretical stability properties. This practically restricts the system to slow, conservative motions, and imposes strong limitations on the dynamics of processes that can be controlled by the end effector. For a detailed treatment of the drawbacks of non-collocated force-control, we refer to~\cite{eppinger1986dynamic, eppinger1987understanding, howard1990joint}.

Moreover, for many of the aforementioned tasks, classical, electrically actuated robot arms without compliant, vibration-damping mechanical elements will not be suited for long runtimes and everyday application. Bad load-cases can rapidly damage sensitive mechanical elements such as gearboxes and will cause them to wear out rapidly. In robotics, common solutions are to consider series-elastic or hydraulic actuators.

Consequently, the next-generation In situ Fabricator needed to be thoroughly rethought in order to provide a concept which resolves these problems. It's worth mentioning that sufficiently-sized platforms with full force-torque control at joint level, and access to low-level control loops (for implementing dynamically capable control methods) are commercially not available today.

\section{Developing the next-generation In situ Fabricator}
\label{sec:next_gen_if}
From the experience gained with IF1 and the Mesh Mould project, it is clear that the next-generation In situ Fabricator (IF2) has to fulfil an additional set of requirements:
\begin{itemize}
\item \textbf{Agility}: Able to perform a specified set of manoeuvres typical of operating in a representative building, for example traverse a narrow doorway from a corridor.
\item \textbf{Payload}: Capable of operating with a \SI{60}{\kg} payload.
\item \textbf{System weight}: Maximum of \SI{440}{\kg} overall system weight (\SI{500}{\kg} including payload), which corresponds to a typical maximum load for a standard floor.
\item \textbf{Arm(s)}: at least one 7~DOF robotic arm with at least \SI{2.5}{\metre} reach.
\item \textbf{Safety}: Capable of reverting to safe or passive modes on detection an of unsafe situation.
\item \textbf{Robust in construction site}: Minimise or eliminate external components that may be subject to damage, e.g. external hoses and wires. Maximise reliability and on-site maintainability.
\item \textbf{Control}: Capable of high bandwidth ($>$\SI{1}{\kHz}) force and position control.
\end{itemize}
Through straight-forward calculations it can be shown that using conventional electrical or hydraulic robot joint actuators, it would be impossible to achieve the required \SI{440}{\kg} overall mass limit and the desired PWR at the same time. An assessment of the other system requirements suggests that a completely novel actuator design is required in order to achieve the desired performance, weight and force control properties. 
\subsection{Intermediate result: development of a novel type of hydraulic actuator}
\label{sec:novel_hydraulic_actuator}
\begin{figure}[tb]
    \centering
    \includegraphics[width=0.98\columnwidth]{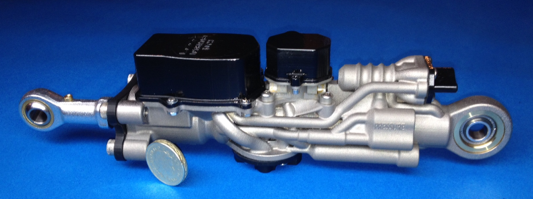}
    \caption{Linear Integrated Actuator, see~\cite{semini2016overview} for details.}    
    \label{fig:linear_integrated_actuator} 
\end{figure}
\begin{figure}[tb]
    \centering
   \includegraphics[width=0.98\columnwidth]{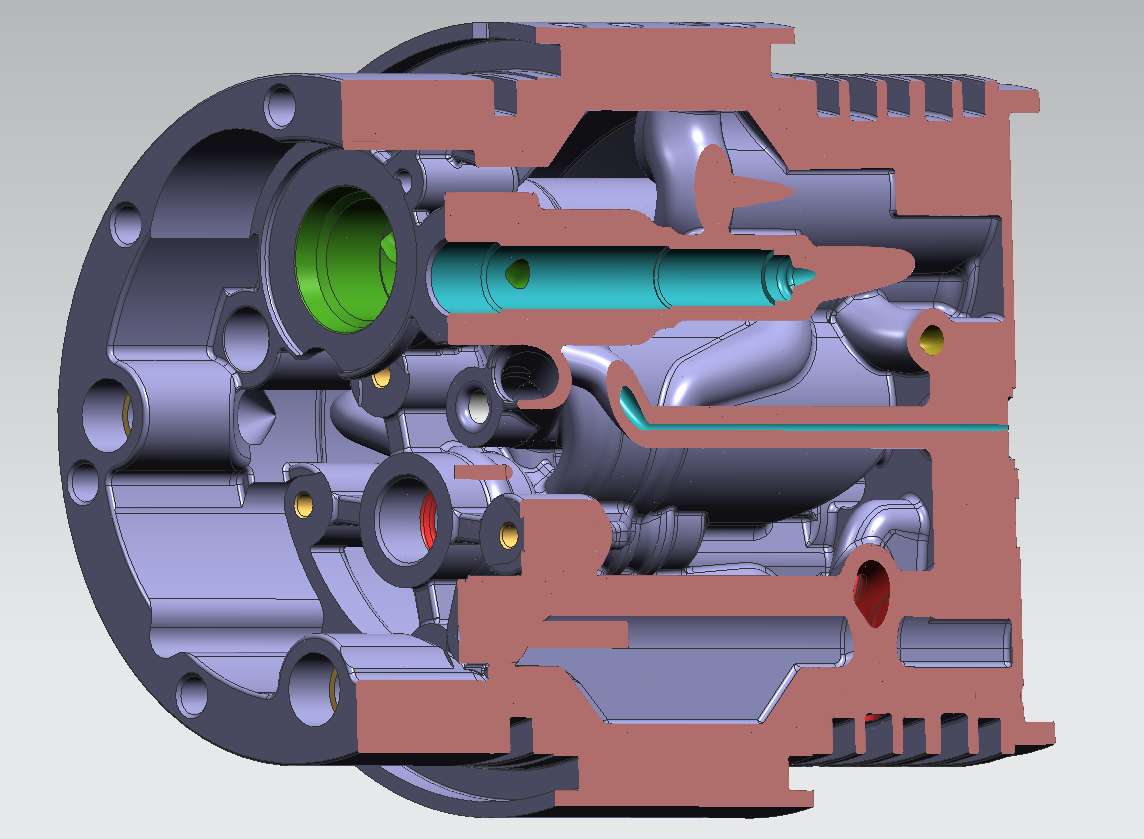}
   \caption{Section through a fully integrated vane actuator as described in Section~\ref{sec:next_gen_if}. It includes all hydraulic controls, sensors, electronics, local processing, data bus and slip rings.}
   \label{fig:actuator_section}
\end{figure}
In order to meet the demanding requirements for IF2, a highly feature dense, structurally capable, lightweight actuator design is required. In cooperation with Moog~Inc and Renishaw~PLC, we are developing a novel, integrated hydraulic actuator offering superior power density. Building on previous work in the field of integrated actuation~(Figure~\ref{fig:linear_integrated_actuator}), a novel titanium fully integrated vane actuator is developed (see Figure~\ref{fig:actuator_section}). The actuator is constructed around a conventional limited angle rotary vane actuator and includes all hydraulic controls, safety valves, sensors, electronic controls, local processing, data bus as well as hydraulic and electrical slip rings. In order to provide the required degree of integration in a compact, structurally efficient package, additive manufacturing using the \emph{laser powder bed} principle was chosen for the major components.

The actuator is capable of mounting for joint rotation around and perpendicular to the major axis. It is designed in three different sizes, which for example allows for a lightweight realisation of different manipulator segments such as arms or legs.

\subsection{Outlook on IF2}
\label{sec:outlook_on_if2}

Hydraulic actuation in conjunction with advanced additive manufacturing technology is particularly well suited to construction robots. It features superior power-density, even on compact, mobile systems with on-board pumps. At the same time, hydraulic systems can be scaled up to dimensions relevant for construction machinery more easily than electrically actuated systems and are typically highly robust -- one of the reasons why a majority of heavy-duty machinery on today's construction site is hydraulically driven.

\begin{figure}[tb]
    \centering
    \includegraphics[width=0.99\columnwidth]{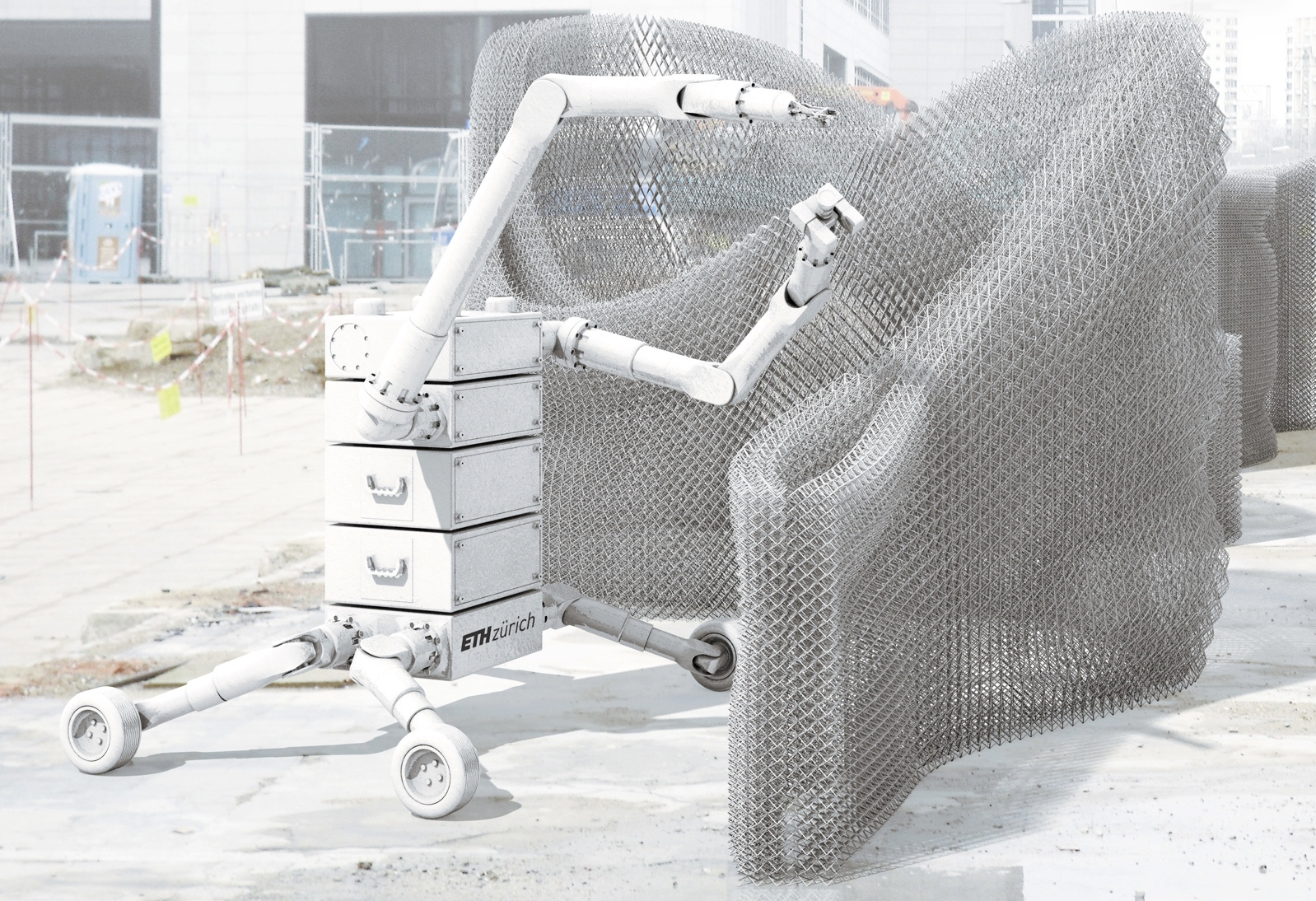}
    \caption{Concept of the In situ Fabricator 2, which is currently under development.}
    \label{fig:if2_rendering}
\end{figure}

Based on our experience with IF1 and the novel hydraulic actuators, a preliminary design of IF2 has been created, which is shown in Figure~\ref{fig:if2_rendering}. In order to achieve the desired manoeuvrability, it is equipped with legs and wheels, which allows for multiple modes of locomotion: walking, driving, or hybrid modes. IF2 is currently under development. A first complete prototype and first results are expected by the end of 2017.

\section{Summary and Conclusion}
\label{sec:conclusion}

In this paper, motivated by the need for digitally controlled mobile robots for on-site manufacturing, assembly and digital fabrication, we have presented an overview of a class of machine that we call `In situ Fabricators'. 
We have listed the core requirements defining that class of robot. We have presented a compact overview of the IF1, which is a prototype system based on classical industrial off-the-shelf components, and its capabilities. In order to meet the desired accuracy and performance, we have implemented a number of state-of-the art algorithms for motion planning, state estimation and control. The development and implementation of the IF1's software framework was strongly inspired by the needs of two full-scale application demonstrators, which are showcased in this paper:
First, a full-size undulating brick wall, which required a number of repositioning manoeuvres during the building process, in which we achieved mm-scale positioning accuracy. Sec\-ond, the Mesh Mould process, which shows that an In situ Fabricator in combination with an innovative toolhead can be used to enable completely novel building processes. IF1 successfully built a number of metal mesh segments, which were also filled with concrete and underwent structural load tests. In a next step, IF1 will be deployed to an exploratory construction site in order to build the ground floor of a demonstrator building.

We also emphasised the limitations of our approach. As the general interest in construction robotics and digital fabrication is currently increasing in both academia and industry, one of our core aims is to raise the awareness amongst other researchers in the field, that the classical industrial robotics approach is bound to a number of significant disadvantages. In our case, these limitations provided the motivation for the development of an innovative, compact, force-controlled rotary hydraulic actuator. Thanks to very recent developments in AM technology we are enabled to use highly integrated compact actuators in conjunction with very efficient additively manufactured structural components. We concluded this work by introducing the concept for IF2, the next-generation In Situ Fabricator. We expect that this development is going to be a major step towards facilitating advances in full-scale construction robots.


\bibliographystyle{spmpsci}      
\small
\bibliography{root}   

%
%

\end{document}